\documentclass[conference]{IEEEtran}
\IEEEoverridecommandlockouts
\usepackage{cite}
\usepackage{amsmath,amssymb,amsfonts}
\usepackage{ascmac}
\usepackage{url}
\usepackage{algorithmic}
\usepackage{graphicx}
\usepackage{textcomp}
\usepackage{xcolor}
\def\BibTeX{{\rm B\kern-.05em{\sc i\kern-.025em b}\kern-.08em
    T\kern-.1667em\lower.7ex\hbox{E}\kern-.125emX}}
\usepackage{multirow}
\renewcommand\IEEEkeywordsname{Keywords}
\usepackage{hyperref}
\hypersetup{
    colorlinks=true,
    citecolor=blue,
    linkcolor=blue,
    urlcolor=blue,
}
\usepackage{tikz}

\newcommand\copyrighttext{%
  \footnotesize \textcopyright 2024 IEEE.  Personal use of this material is permitted.  Permission from IEEE must be obtained for all other uses, in any current or future media, including reprinting/republishing this material for advertising or promotional purposes, creating new collective works, for resale or redistribution to servers or lists, or reuse of any copyrighted component of this work in other works.}
\newcommand\copyrightnotice{%
\begin{tikzpicture}[remember picture,overlay]
\node[anchor=south,yshift=10pt] at (current page.south) {\fbox{\parbox{\dimexpr\textwidth-\fboxsep-\fboxrule\relax}{\copyrighttext}}};
\end{tikzpicture}%
}

\makeatletter 
\newcommand{\linebreakand}{%
  \end{@IEEEauthorhalign}
  \hfill\mbox{}\par
  \mbox{}\hfill\begin{@IEEEauthorhalign}
}
\makeatother 

\begin{document}

\title{Coarse-Grained Sense Inventories Based on Semantic Matching between English Dictionaries}

\author{\IEEEauthorblockN{Masato Kikuchi}
\IEEEauthorblockA{\textit{Nagoya Institute of Technology}\\
Nagoya, Aichi, Japan \\
kikuchi@nitech.ac.jp}
\and
\IEEEauthorblockN{Masatsugu Ono}
\IEEEauthorblockA{\textit{Muroran Institute of Technology}\\
Muroran, Hokkaido, Japan \\
onomasa@muroran-it.ac.jp}
\and
\IEEEauthorblockN{Toshioki Soga}
\IEEEauthorblockA{\textit{Chitose Institute of Science and}\\
\textit{Technology}\\
Chitose, Hokkaido, Japan \\
t-soga@photon.chitose.ac.jp}
\and 
\linebreakand
\IEEEauthorblockN{Tetsu Tanabe}
\IEEEauthorblockA{\textit{Hokkaido University}\\
Sapporo, Hokkaido, Japan \\
ttanabe@iic.hokudai.ac.jp}
\and 
\IEEEauthorblockN{Tadachika Ozono}
\IEEEauthorblockA{\textit{Nagoya Institute of Technology}\\
Nagoya, Aichi, Japan \\
ozono@nitech.ac.jp}
}

\maketitle
\copyrightnotice{}

\begin{abstract}

WordNet is one of the largest handcrafted concept dictionaries visualizing word connections through semantic relationships. It is widely used as a word sense inventory in natural language processing tasks. However, WordNet's fine-grained senses have been criticized for limiting its usability. In this paper, we semantically match sense definitions from Cambridge dictionaries and WordNet and develop new coarse-grained sense inventories. We verify the effectiveness of our inventories by comparing their semantic coherences with that of Coarse Sense Inventory. The advantages of the proposed inventories include their low dependency on large-scale resources, better aggregation of closely related senses, CEFR-level assignments, and ease of expansion and improvement. Our inventories are publicly available for free use\footnote{Our inventories are publicly available at \url{https://doi.org/10.5281/zenodo.13706831}.}.

\end{abstract}


\section{Introduction}

WordNet\footnote{\url{https://wordnet.princeton.edu/} (accessed: July 1, 2024)}~\cite{Fellbaum:98} is one of the largest handcrafted, freely available concept dictionaries. It visualizes connections between English words through semantic concepts such as hypernyms, hyponyms, and synonyms. WordNet is often used for natural language processing (NLP) tasks, because it provides a machine-friendly word sense inventory. However, WordNet has been criticized for its overly fine-grained word senses.
We introduce an NLP task, called word sense disambiguation (WSD), as an example. WSD involves identifying the correct sense of a word based on its context. WSD systems use sense inventories to train their models and output the appropriate senses. Therefore, the choice of inventory affects the performance of the WSD system, and the fine-grained senses of WordNet often lead to poor performance.
Fig.~\ref{fig:wsd_ex} uses two sense candidates—which are indeterminable even to humans—to illustrate the challenging problems WSD systems must tackle.

\begin{figure}
\begin{screen}
I stopped to \textbf{say} goodbye, Mrs. Lattimer, and to tell you how sorry I was to hear about your baby.\\\\
Candidates: (1) utter aloud $\quad$ (2) express in words
\end{screen}
\caption{An example of WSD. The target word ``say'' for WSD is in bold. The correct sense is assigned to (1), but (2) could also be considered correct. However, the system must choose only one sense.}
\label{fig:wsd_ex}
\end{figure}
We are currently working on the `Learner's WordNet Project,' which aims to build a novel WordNet particularly for second language learners. The significance of this project lies in incorporating the excellent semantic concepts of WordNet into English language learning. However, WordNet's senses are created from a linguistic perspective, significantly different from the coarse-grained level used in English learning textbooks. This makes it challenging to use WordNet in education.
Therefore, we propose grouping WordNet's senses based on the senses in Cambridge dictionaries\footnote{\url{https://dictionary.cambridge.org/us/} (accessed: July 1, 2024)}, commonly used in education, to create new coarse-grained sense inventories. We verify the effectiveness of our inventories by comparing the coherences of the sense groups with that of the Coarse Sense Inventory (CSI)~\cite{Lacerra:20}.
The advantages of the proposed inventories are as follows:
i) Because our inventories are created using only sense definitions based on select dictionaries, they do not require large-scale external resources.
ii) The inventory creation process is fully automated, making it easy to expand and improve.
iii) Compared with CSI, which is created manually, our approach successfully aggregates senses that are very close in meaning.
iv) The groups are assigned CEFR levels, indicating their priority in language learning.

\section{Related Work}

One criticism of WordNet is that its senses are excessively fine-grained, making it difficult even for humans to distinguish among them. In this section, we review the coarse-grained inventories developed to overcome this problem. Additionally, we review studies on sentence similarity measures closely related to our research.

According to Izquierdo et al.~\cite{Izquierdo:15}, sense inventories are classified into word- and class-based inventories. Most word-based approaches attempt to cluster WordNet senses associated with the same word, thereby addressing the problem of fine granularity~\cite{Palmer:04, Hovy:06, Navigli:06, Palmer:07, Snow:07}. Among them, Navigli~\cite{Navigli:06} clustered the WordNet inventory and mapped its senses to the Oxford English Dictionary. The author proposed a methodology similar to ours, mapping WordNet senses to another English dictionary, which serves as one justification for our approach. The problem with these approaches is that they require many sense-tagged usage examples for clustering. These approaches miss many rare senses, because the frequency distribution of word senses follows a power law.

Class-based approaches mitigate the problem of rare senses by providing labels shared among various words. CSI, introduced in Section~\ref{sec:csi}, also belongs to this category. Many class-based inventories have coarse-grained labels, which are easy to interpret semantically. These inventories refer to a variety of knowledge, such as WordNet's lexicographer files, WordNet relations~\cite{Izquierdo:07}, hypernymy~\cite{Vial:19}, WordNet Domains~\cite{Magnini:00}, and external sources~\cite{Camacho-Collados:17, Lacerra:20} to define labels with good interpretability, mapping one or more senses in WordNet. However, the number of label types is limited compared with the diversity of word meanings, making the inventories coarser than word-based inventories. Highly coarse-grained inventories are less valuable because they fail to maintain a meaningful granularity for humans and computers.

Our inventories do not require usage examples of word senses and can be created using only sense definitions. In other words, they can cover rare senses that conventional word-based inventories cannot handle. Additionally, the inventories can suppress the inclusion of inappropriate senses, which can occur in class-based inventories, in sense groups.
Coarse-grained sense inventories may also have potential applications in foreign language learning. Aiba~\cite{Aiba:2021} investigated how sense groups matched in NLP-based and second language learners' ability-based lexical resources (OntoNotes~\cite{Hovy:06} and English Vocabulary Profile, respectively). Although the study was small, involving 47 randomly selected verbs, the author found that the groups overlapped by 81.6\%, suggesting that both resources may be applicable for two different purposes: NLP and language learning.
Our resources are annotated with CEFR levels, the international foreign language proficiency index, further enhancing their applicability in language learning.

Measuring text similarity is a primary challenge in this study. Many sentence similarity measures have been proposed to tackle this challenge~\cite{Chandrasekaran:21, Prakoso:21}. These techniques are used in various NLP tasks, such as information retrieval, automatic answering of questions, machine translation, dialogue systems, and document matching. However, few studies have applied similarity measures to sense definitions across dictionaries to create coarse-grained sense inventories.
Yao et al.~\cite{Yao:21} proposed an alignment algorithm for sense definitions to integrate vocabulary knowledge from multiple sense inventories. However, the objective of their study was to enhance semantic equivalence in WSD, particularly for rare word senses, using the similarity measures, which is different from the objective of the present study.
The sense definitions we measured similarity for have the following characteristics:
First, even if the compared senses are similar, the length of sentences and the words used in the definitions may differ significantly.
Second, there are almost no cases where the senses are entirely equivalent; therefore, it is necessary to handle cases where the definitions are slightly different flexibly.
We used a large language model (LLM) to perform prompt-based similarity measurements based on these characteristics. The LLM facilitates replicating our approach and establishes a flexible measure of dealing with similarities by adjusting prompts.

\section{Coarse-grained Sense Inventories}

First, we discuss the existing sense inventory CSI. Next, we introduce our inventories and their construction method and structure.

\subsection{CSI}
\label{sec:csi}

To the best of our knowledge, the most recent existing inventory is CSI. This inventory is class-based, with coarse-grained labels shared between various words. CSI was created in the following steps: i) categories from a well-known thesaurus, \textit{Roget Thesaurus}, were manually clustered to define the inventory, and ii) annotators mapped WordNet synsets to one or more CSI labels.
This inventory has two advantages. First, the coarse-grained sense labels have excellent interpretability. For example, labels LAW\&CRIME and OLFACTORY provide a rough understanding of the sense group's meaning just by seeing them. Second, the inventory has a high coverage rate for WordNet synsets. At least one CSI label is assigned to 83K synsets, resulting in a coverage rate of 70.4\%.
However, CSI has two disadvantages. First, it is difficult to scale up because many parts of the creation process are manual. Second, the mapping of sense keys to labels is highly coarse-grained. The number of CSI labels is limited compared with the diversity of possible word meanings; therefore, the mapping may be excessively coarse-grained for WSD, potentially grouping dissimilar senses.

\subsection{Proposed Inventories}

\begin{figure}
\hrulefill

Please assess whether the two meanings of the English word `WORD' are the same from a linguistic perspective.\\
1: [one sense definition of `WORD' in CLD or CED]\\
2: [one sense definition of `WORD' in WordNet]\\
\\
Please select one from the following and answer using only number.\\
1. Exactly the same meaning\\
2. Almost the same meaning\\
3. Somewhat similar meaning\\
4. Neither similar nor different meaning\\
5. Somewhat different meaning\\
6. Mostly different meaning\\
7. Completely different meaning

\hrulefill
\caption{Prompt template used for word sense matching}
\label{fig:prompt_ex1}
\end{figure}
\begin{figure*}[tb]
  \centering
  \includegraphics[keepaspectratio, scale=0.4]{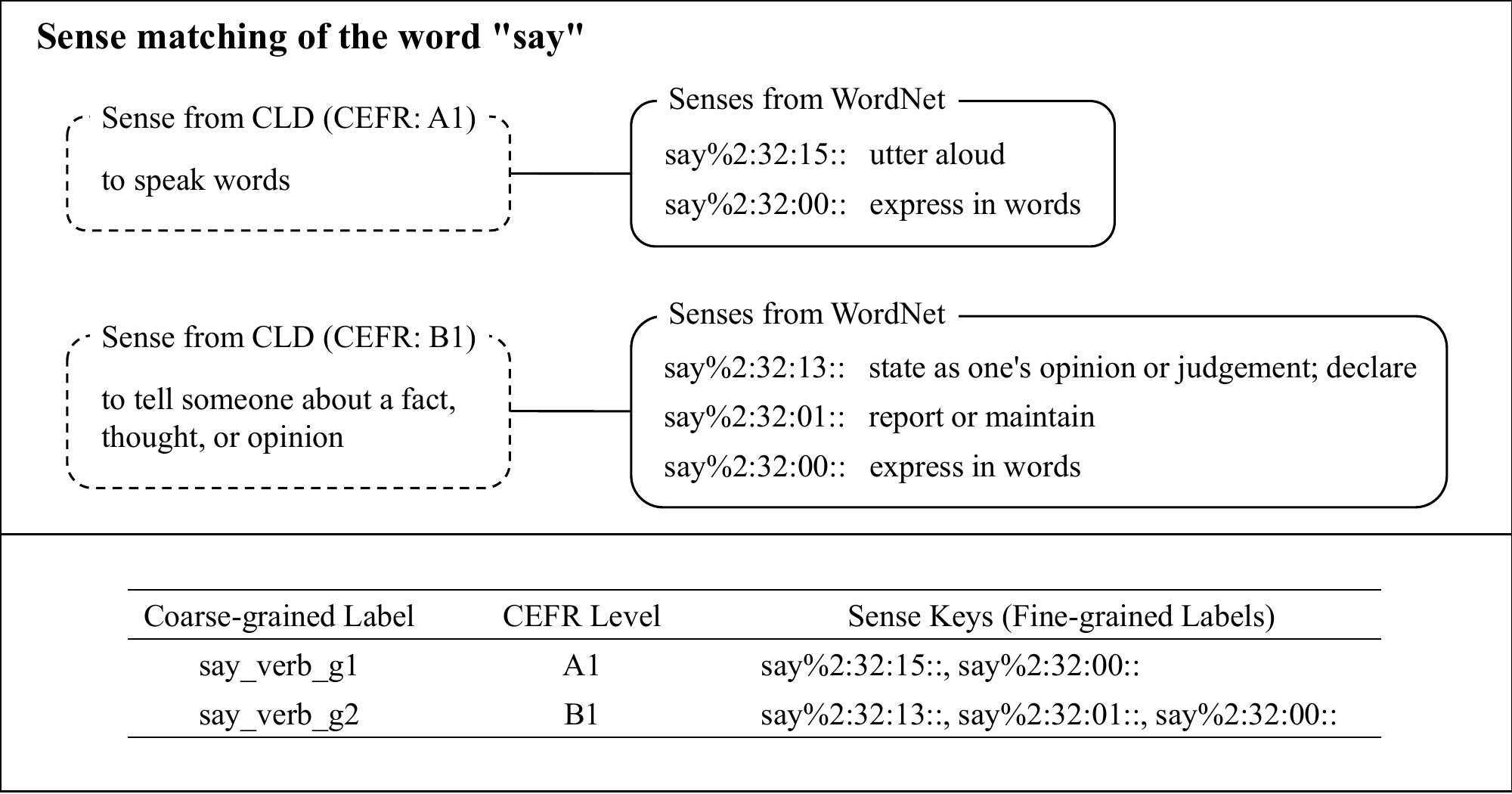}
  \caption{Sense matching examples (top) and coarse-grained sense representations (bottom) constructed using them}
  \label{fig:matching_ex}
\end{figure*}
We created two sense inventories based on sense matching between Cambridge dictionaries and WordNet: one based on semantic matching between Cambridge Learner's Dictionary (CLD) and WordNet, and the other based on semantic matching between the Cambridge English Dictionary (CED) and WordNet. The inventories comprised 3,222 and 9,457 sense groups, respectively. These inventories were created as follows:
First, we selected 15,885 English words listed in various vocabulary lists\footnote{The vocabulary lists included COCET3300, Sugiura English List I, Sugiura English List II, BNC Vocabulary List, Hokkaido
University English Vocabulary List, JACET8000, and ALC-SVL12000.} used in English learning in Japan as target words. Commonly used words tend to have multiple senses~\cite{Zipf:45}, and many selected words have numerous fine-grained senses. Next, we extracted sense information (part of speech (PoS), sense key, and sense definition) from WordNet for each target word.
Next, we comprehensively compared the extracted sense definitions with those from Cambridge dictionaries to determine the match of senses. We used ChatGPT (gpt-4o-2024-05-13) to perform this, ensuring only those with the same PoS were compared to avoid unnecessary comparisons. ChatGPT received the prompt shown in Fig.~\ref{fig:prompt_ex1}. We set the system role context to "You are a professional linguist." and the ChatGPT parameter temperature to zero.
To allow for flexible semantic matching, we defined the degree of matching on a seven-point scale, as shown in Fig.~\ref{fig:prompt_ex1}, and considered sense pairs with outputs of 1 and 2 to be a match. Finally, we extracted combinations in which multiple WordNet senses matched one sense in each Cambridge dictionary and grouped them into a single coarse-grained sense.

Fig.~\ref{fig:matching_ex} shows examples of sense matching and the resulting coarse-grained sense representations. In this figure, the sense of ``say'' in CLD, ``to speak words,'' matches two senses in WordNet: ``utter aloud'' and ``express in words.'' The identifiers say\%2:32:15:: and say\%2:32:00:: are called sense keys, representing specific senses in WordNet. Similarly, another sense in CLD matches the three senses in WordNet. The sense representations are constructed by grouping sense keys based on these matches.
Although our coarse-grained labels cannot infer semantic relevance like CSI, our inventories has the following advantages. First, the creation process is fully automated, and our inventories can be flexibly expanded without manual effort. Second, the criteria for grouping senses are transparent, allowing even rare senses to be grouped effectively. Our inventories can be constructed without sense-tagged example sentences in a corpus, making them independent of the frequency of senses in the corpus. Third, CEFR levels are assigned to the coarse-grained senses. The CEFR level is an international index indicating foreign language proficiency, enabling one to gauge the difficulty of grouped senses.
To the best of our knowledge, no other inventory assigns CEFR levels to coarse-grained labels. Therefore, our inventories are particularly useful for English learning applications.

\section{Experiments}

We performed experiments to evaluate the quality of coarse-grained inventories by assessing the ``cohesiveness'' of the sense groups included in each inventory. Herein, a group is considered cohesive if it accurately and completely covers only similar senses. A high-quality sense inventory consists of well-formed, cohesive sense groups.
In our experiments, we used the LLM (ChatGPT) to distinguish between senses within the same group. Assuming closely related senses might confuse even humans, groups in which the LLM fails to distinguish are considered better. Additionally, to confirm comprehensive coverage of each group, the LLM attempts to distinguish between senses belonging to a group and those not belonging to the group. If groups comprehensively include all similar senses, it should perfectly distinguish between the senses.

\subsection{Procedure and Settings}

\begin{figure}
\hrulefill

Is the word `WORD,' which is enclosed in \#\#\#, used in the following sentence with the following meaning?\\
Please answer YES or NO.\\
Sentence: [A sentence tagged with the sense key of $s$]\\
Meaning: [The sense definition of $s^{\prime}$]

\hrulefill

Is the word `say,' which is enclosed in \#\#\#, used in the following sentence with the following meaning?\\
Please answer YES or NO.\\
Sentence: I stopped to \#\#\#say\#\#\# goodbye, Mrs. Lattimer, and to tell you how sorry I was to hear about your baby.\\
Meaning: express in words

\hrulefill
\caption{Template (top) and example (bottom) of the prompt used in our experiments. In this example, WORD is ``say,'' and the sense definitions of $s$ and $s^{\prime}$ are ``utter aloud'' and ``express in words,'' respectively.}
\label{fig:prompt}
\end{figure}
Let $I$ be a coarse-grained sense inventory and $S_w(i)\in I$ be a sense group in $I$. $S_w(i)$ indicates the $i$-th sense group of word $w$. In the first experiment, we confirmed if ChatGPT confuses two senses $s,\ s^{\prime} \in S_w(i)$. The procedure was as follows:
First, we randomly sampled 1,000 sense groups from $I$. Next, we randomly selected $s$ and $s^{\prime}$ from each group and embedded them in the prompt template in Fig.~\ref{fig:prompt}. We obtained the sentence using $s$ by randomly selecting a sentence annotated with the sense key of $s$ from the sense-tagged corpus SemCor 3.0. We obtained the sense definition of $s^{\prime}$ from WordNet. Consequently, we created 1,000 prompts.
Finally, we input each prompt to ChatGPT and collected the responses (YES or NO). To ensure randomness, we repeated the above procedure five times and averaged the aggregated results.

In the second experiment, we confirmed the ability of ChatGPT to distinguish between the senses $s \in S_w(i)$ and $s^{\prime} \in \bar{S}_w(i)$ for the word $w$. Here, $\bar{S}_w(i)$ represented the exclusive set of $S_w(i)$, i.e., the set of senses of $w$ not included in $S_w(i)$. The procedure was as follows:
First, we randomly sampled 1,000 sense groups from $I$. Next, we extracted $s$ from $S_w(i)$ and $s^{\prime}$ from $\bar{S}_w(i)$, and embedded them in the template in Fig.~\ref{fig:prompt}. We obtained the sentence for $s$ and the sense definition for $s^{\prime}$ using the same methods as in the first experiment. Consequently, we created 1,000 prompts.
Finally, we input each prompt to ChatGPT and collected the responses. To ensure randomness, we repeated this procedure five times and averaged the aggregated results.

We used CSI and our inventories based on CLD and CED for comparison. For a fair comparison, we used only those senses in CSI that corresponded to the words in our inventories. A sense group in CSI sometimes contains senses with different PoSs. However, when attempting WSD, the PoS is predetermined. Therefore, we selected $s$ and $s^{\prime}$ with the same PoS for the experiments.
We used ChatGPT version gpt-4o-2024-05-13, the latest version at the time of the experiment. We set the system role context to ``You are a professional linguist.'' and the ChatGPT parameter temperature to zero.

\subsection{Experimental Results}
\label{sec:results}

Before conducting our experiments, we evaluated ChatGPT's ability to recognize senses. We randomly extracted 1,000 senses from WordNet and created prompts as shown in Fig.~\ref{fig:prompt}, where $s$ and $s^{\prime}$ are identical, and input them to ChatGPT. If ChatGPT fully recognizes the senses, all responses should be YES. The result showed that the proportion of YES responses was 825 out of 1,000, indicating that ChatGPT correctly recognizes over 80\% of the senses.

\begin{table}[tb]
\caption{Results of Experiment 1}
\label{tab:results1}
\centering
	\begin{tabular}{ c  r  r  c } \hline
	\multicolumn{1}{ c }{\multirow{2}{*}{Inventory}} & \multicolumn{2}{ c }{Average} & \multicolumn{1}{ c }{\multirow{2}{*}{Ratio of YES ($\uparrow$)}} \\ \cline{2-3}
		& \multicolumn{1}{ c }{YES} & \multicolumn{1}{ c }{NO} & \\ \hline
		CSI & 388 & 612 & 0.388\\
		CLD (Ours) & 675 & 325 & 0.675 \\
		CED (Ours) & 656 & 344 & 0.656 \\ \hline
	\end{tabular}
\end{table}
Experiment 1 involved identifying probably very similar but different senses. When an inventory aggregates confusing senses that challenge even humans, the number of YES responses should be large because ChatGPT cannot correctly identify many senses, even though all answers should be NO. TABLE~\ref{tab:results1} lists the results of the experiment.
For CSI, YES responses were less than 40\%, indicating that the group contained many distinguishable senses for ChatGPT. This result is undesirable, because the objective of creating CSI has been to improve the performance of WSD by grouping confusing senses that humans cannot distinguish. In contrast, YES responses were over 65\% for our two inventories, indicating that ChatGPT was more likely to misidentify senses. This result suggests that our inventories can aggregate confusing senses better than does the CSI.

\begin{table}[tb]
\caption{Results of Experiment 2}
\label{tab:results2}
\centering
	\begin{tabular}{ c  r  r  c } \hline
	\multicolumn{1}{ c }{\multirow{2}{*}{Inventory}} & \multicolumn{2}{ c }{Average} & \multicolumn{1}{ c }{\multirow{2}{*}{Ratio of NO ($\uparrow$)}} \\ \cline{2-3}
		& \multicolumn{1}{ c }{YES} & \multicolumn{1}{ c }{NO} & \\ \hline
		CSI & 166 & 834 & 0.834 \\
		CLD (Ours) & 180 & 820 & 0.820 \\
		CED (Ours) & 151 & 849 & 0.849 \\ \hline
	\end{tabular}
\end{table}
Experiment 2 involved identifying senses within and beyond a group. If an inventory aggregates confusing senses without missing any, the senses outside the group will not be similar to those in the group. Therefore, ChatGPT should correctly answer NO to all prompts. TABLE~\ref{tab:results2} lists the results of the experiment.
The percentage of YES responses was less than 20\% for all inventories, with our CLD-based inventory having the highest percentage of YES. This is because CLD, used for grouping, does not cover senses as comprehensively as WordNet, leading to the possibility that some ungrouped senses are similar to those in the group. Unlike in Experiment 1, CSI shows a tendency similar to our inventories.
Considering this and the results of Experiment 1, we infer that CSI contains an excessive number of senses in each group. When similar senses are included in the group without exception, and other unnecessary senses are also included, suitable results may only appear in Experiment 2. In contrast, although the CLD-based inventory was slightly inferior to CSI, the difference was not as significant as in Experiment 1, suggesting that our inventories aggregated similar senses without including unrelated senses.

\section{Discussion}

\begin{figure}
\begin{screen}
CLD: to discover something or someone that you have been searching for\\
$s$: come upon after searching; find the location of something that was missed or lost\\
$s^{\prime}$: make a discovery
\end{screen}
\caption{Example showing one sense definition in CLD matching two WordNet definitions with different meanings. All definitions are of the word ``find.''}
\label{fig:failure_ex}
\end{figure}
The results of our experiments clarified that our inventories aggregated similar senses better than did the CSI. However, in Experiment 1, we found that ChatGPT could correctly distinguish approximately 35\% of the sense pairs. We are of the view that the creation process of our inventories is related to why distinct senses were mixed into the groups, and we qualitatively analyzed this issue.
For prompts to which ChatGPT correctly answered NO, we visually examined the sense definitions of $s$ and $s^{\prime}$ and the definitions in Cambridge dictionaries. We found that although the matching itself was correct, various senses were sometimes mixed into the same group. As shown in Fig.~\ref{fig:failure_ex}, one sense definition of CLD is divided into two sense definitions in WordNet. In this case, the sense of CLD matched both $s$ and $s^{\prime}$ as ``Almost the same meaning,'' but their definitions varied. We consider that such differences in sense definitions are one of the factors that produced the experimental results.
By using only groups based on strict matching, i.e., groups of senses that match ``Exactly the same meaning,'' it is possible to create an inventory that aggregates senses with a higher degree of semantic agreement. However, senses that are subtly different in expression but often used in common situations, such as ``to speak words'' and ``express in words'' for the word ``say,'' may be excluded from the group, even though these senses should be aggregated into the same group. Another problem is that an inventory based on strict matching contains only a few senses. Therefore, there is scope for improvement in the creation process for coarse-grained sense inventories.

\section{Conclusion}

WordNet is an electronic sense inventory used in NLP and language learning; however, its word senses are extremely fine-grained. Herein, we proposed new coarse-grained sense inventories by aligning the sense definitions given in Cambridge dictionaries and WordNet. Our resources have the advantages of an automatic creation process and not relying on large-scale external resources. We compared the inventories with CSI and confirmed that the proposed inventories appropriately aggregated similar senses. The proposed inventories are publicly available for free use\footnotemark[1]. Our coarse-grained sense groups were assigned CEFR levels, which are expected to be applied to foreign language learning. We plan to evaluate the assigned CEFR levels and summarize the results in a separate paper. Additionally, we will explore how fine-tuning the LLMs and prompt engineering can produce higher-quality inventories.

\section*{Acknowledgment}

This work was supported in part by JSPS KAKENHI Grant Numbers JP22K02825 and JP22K18006.

\section*{Appendix}

\begin{figure}[tb]
  \centering
  \includegraphics[keepaspectratio, scale=0.65]{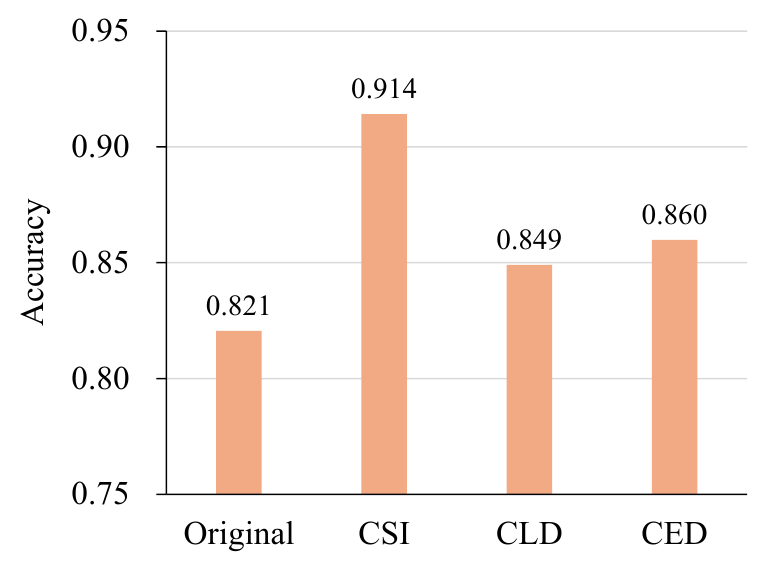}
  \caption{WSD performance based on each inventory}
  \label{fig:WSD_eval}
\end{figure}
We utilized coarse-grained inventories to explore the actual performance of the state-of-the-art WSD model. As a WSD dataset, we used the ALL dataset\footnote{\url{http://lcl.uniroma1.it/wsdeval/evaluation-data} (accessed: July 25, 2024)}~\cite{Raganato:17}, which is a combination of datasets used in the WSD competitions Senseval and SemEval. This dataset contains 7,253 target words and their contexts (sentences). We used only the target words in the vocabulary lists\footnotemark[4]. Of the 7,253 words, 6,605 were retained, indicating that our inventories contained many words subject to WSD.
We used CONtinuous SEnse Comprehension (ConSeC)~\cite{Barba:21} as the WSD model, which achieved a high performance of 83.2\% on the all-words WSD task using fine-grained senses. We applied each coarse-grained inventory in the following way: First, we ran ConSeC for fine-grained WSD and extracted the incorrectly predicted sense keys. Then, we referred to the inventory and checked whether the keys were in the same group as the correct keys. When they did, we treated the keys as correct.

Fig.~\ref{fig:WSD_eval} shows the evaluation results. In this evaluation, we used only target words listed in the vocabulary lists. Therefore, the original performance of ConSeC dropped slightly from 83.2\% to 82.1\%. It is evident that the CSI result is significantly better than the original result. However, this may overestimate performance. When using our CLD/CED--WordNet inventories, the performance of ConSeC increased slightly to approximately 85\% and 86\%, respectively. Therefore, considering our experimental results in Section~\ref{sec:results}, we conclude that the actual performance of ConSeC is approximately 85-86\%.

\bibliographystyle{IEEEtran}
\bibliography{references}

\end{document}